\documentclass[conference]{IEEEtran}
\IEEEoverridecommandlockouts

\usepackage{cite}
\usepackage{amsmath,amssymb,amsfonts}
\usepackage{algorithmic}
\usepackage{graphicx}
\usepackage{textcomp}
\usepackage{xcolor}
\usepackage{booktabs}
\usepackage{subfigure}
\usepackage{subcaption}

\def\BibTeX{{\rm B\kern-.05em{\sc i\kern-.025em b}\kern-.08em
    T\kern-.1667em\lower.7ex\hbox{E}\kern-.125emX}}
\begin{document}

\title{Estimation of Minimum Stride Frequency for the Frontal Plane Stability of Bipedal Systems}

\author{\IEEEauthorblockN{Harsha Karunanayaka}
\IEEEauthorblockA{\textit{Department of Mechanical Engineering} \\
\textit{University of Denver}\\
Denver, CO, USA\\
harsha.karunanayakamudiyanselage@du.edu}
\and
\IEEEauthorblockN{Siavash Rezazadeh}
\IEEEauthorblockA{\textit{Department of Mechanical Engineering} \\
\textit{University of Denver}\\
Denver, CO, USA\\
Siavash.Rezazadeh@du.edu}
}

\maketitle

\begin{abstract}
Stability of bipedal systems in frontal plane is affected by the hip offset, to the extent that adjusting stride time using feedforward retraction and extension of the legs can lead to stable oscillations without feedback control. This feedforward stabilization can be leveraged to reduce the control effort and energy expenditure and increase the locomotion robustness. However, there is limited understanding of how key parameters, such as mass, stiffness, leg length, and hip width, affect stability and the minimum stride frequency needed to maintain it. This study aims to address these gaps through analyzing how individual model parameters and the system's natural frequency influence the minimum stride frequency required to maintain a stable cycle. We propose a method to predict the minimum stride frequency, and compare the predicted stride frequencies with actual values for randomly generated models. The findings of this work provide a better understanding of the frontal plane stability mechanisms and how feedforward stabilization can be leveraged to reduce the control effort.
\end{abstract}


\section{Introduction}

The stability of bipedal locomotion depends on maintaining balance in both the sagittal and frontal planes. Although most existing research focuses on the sagittal plane, the mechanisms underlying frontal plane stability remain less explored. In human bipedal locomotion, loss of balance in the medial-lateral plane frequently results in falls, and stability in this plane is influenced by body posture and foot placement \cite{mille2005age}. With the primary aim of bipedal robots being able to achieve human-like locomotion, capturing these mechanisms in robotic systems remains a significant challenge.

\begin{figure*}
	\centering\includegraphics[width=0.9\linewidth]{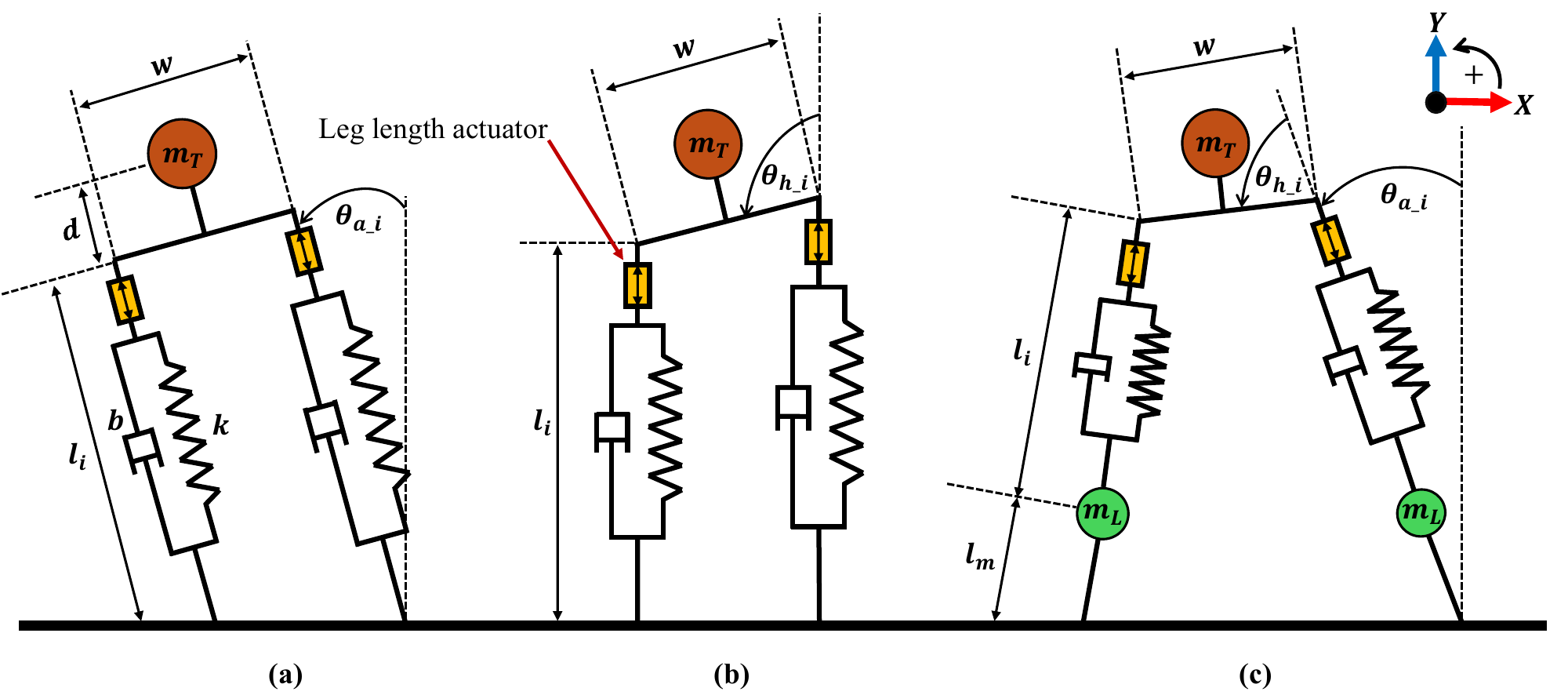}
	\caption{Schematics of the models used; (a) fixed hip, (b) fixed ankle, and (c) both ankle and hip are free to move. The model parameters and variables: $m_T, m_L,k,b,w,d,l_i,l_m,\theta_{a_i},\theta_{h_i}$ denote torso mass, leg mass, leg stiffness, leg damping, hip width, offset of torso mass from hip center, offset of leg mass from the foot, angle of the ankle joint, and angle of the hip joint, respectively. Index $i$ in the variables refers to the leg index (left or right).}
	\label{SLIP_model}
\end{figure*}

The Zero Moment Point (ZMP) is one of the most established strategies for achieving stable locomotion in bipedal and humanoid robots, defined as the point on the ground where the net moment of inertial and gravitational forces is zero. Stability is ensured by maintaining the ZMP within the support polygon, with reference trajectories typically generated using a walking pattern generator and the center of mass (CoM) state. Kajita et al. \cite{kajita2019position} developed a position-based controller for lateral stability in a knee-stretched biped robot by combining ZMP modification and additional CoM acceleration with a Linear Quadratic Regulator (LQR) feedback controller. Similarly, Ito et al. \cite{ito2008zmp} proposed a ZMP feedback controller for in-place lateral stepping. While this method demonstrated promising results in simulation, experimental trials revealed limitations due to slow response and mechanical constraints.

In addition to ZMP-based methods, swing-leg dynamics have also been considered in the locomotion of bipedal robots, as the angular momentum generated by the leg mass is non-negligible. Seyde et al. \cite{8461140} addressed this by presenting a reference trajectory generator based on the Instantaneous Capture Point (ICP) framework extended with angular momentum. Its performance in lateral stepping was evaluated in ATLAS simulations, where it demonstrated improved balance recovery compared with alternative strategies, including the continuous double support (CDS) planner, the continuous heel-to-toe shift (HT) planner, and the smooth ICP planner. In the works of Morimoto et al. \cite{7de9e7e7aa784c1e8f9fa707179496b7}, a biologically inspired biped locomotion strategy was proposed in which the Center of Pressure (CoP) location and velocity were used to define the phases of sinusoidal joint trajectories generated by a coupled oscillator model.

In human locomotion, lateral balance is critical for maintaining the center of mass (CoM) within the narrow base of support defined by the feet. Experimental studies have demonstrated that passive walking is unstable in the frontal plane, necessitating active feedback control through adjustments in foot placement. Although these adjustments are relatively small, they incur a measurable metabolic cost \cite{ijmker2013energy,donelan2004mechanical}. These studies further show that when external lateral stabilization is provided, humans take advantage of reduced step-to-step transition costs at narrower step widths. However, even under external stabilization, higher metabolic costs persist at very narrow step widths, preventing the preferred step width during human walking from approaching zero. Narrower step widths also increase instability in the frontal plane, as the base of support shrinks and any displacement of the CoM outside this base requires larger compensatory motions to preserve stability. In addition, very narrow step widths may demand additional metabolic effort for the lateral displacement of the swing leg away from the stance leg to avoid leg collision, further contributing to the overall energetic cost.

Using a similar experimental approach, Koopman et al. \cite{koopman2013lateral} developed an algorithm to control pelvic motion and provide lateral support during walking, thereby assisting balance rehabilitation in neurological patients. Unlike methods that attract or constrain subjects to the center of the treadmill, this algorithm preserves the natural pelvic sway pattern, thereby supporting balance control in a more physiological manner. Bauby and Kuo \cite{bauby2000active} further demonstrated that lateral balance relies on visual-vestibular feedback and that wider step widths provide a slight advantage for stability. However, Perry and Srinivasan \cite{perry2017walking} reported that wider step widths do not improve linear stability but instead alter foot placement control and increase kinematic variability.

Despite the aforementioned studies on lateral balance and control in walking, limited attention has been given to how model parameters influence lateral stability. In his study of lateral motion in passive walking, Kuo \cite{kuo1999stabilization} reported that increasing step width, spring stiffness, and pelvis radius of gyration, while decreasing the leg radius of gyration, leads to a slight improvement in lateral stability. However, a clear relationship between these parameters and lateral stability has not been established, leaving gaps in understanding how model properties directly affect stability outcomes. Sullivan and Seipel \cite{sullivan20143d} experimentally demonstrated that as step width decreases, variability in roll increases, which indicates reduced roll stability. They consequently hypothesized that dimensionless roll inertia, defined as the ratio of roll inertia to the CoM moment of inertia (as a point mass) around the supporting leg, can serve as a key parameter for understanding lateral stability in bipedal running. Values less than one for this parameter would indicate feedforward stability, whereas values greater than one would require active control to maintain balance. However, these results are based on a single experimental set of results limiting the generalizability of the hypothesis. Moreover, the experiment was conducted at merely one velocity, leaving open the question of how stepping frequency influences lateral stability.

In this study, we investigate how model parameters affect feedforward lateral stability in bipedal locomotion. Our objectives are threefold: (i) to examine the role of both individual parameters and their interactions in shaping variations in the minimum stride frequency, (ii) to determine the minimum stride frequency required for lateral stability without active feedback control, and (iii) to develop a generalized framework for predicting lateral stability across different combinations of model parameters. These contributions provide a foundation for a more systematic understanding of stability mechanisms in bipedal walking.

The organization of this paper is as follows. First, we introduce our mathematical models using compliant legs and periodic leg retraction and extension. Next, we present simulation results for these model and analyze both individual parameters and combinations of them. Based on these results, we then propose a formulation for lateral stability. Finally, we discuss how model parameters influence lateral stability, how the results align with findings from empirical studies on human locomotion, and highlight the limitations and potential future directions for this research.

\section{Methods}
\subsection{Models Used for the Simulations}

We developed a mathematical bipedal model for the frontal plane based on the forced-oscillation extension of the bipedal spring-loaded inverted pendulum (SLIP) framework, which includes a spring, a damping element, and a leg length actuator for each leg \cite{rezazadeh2020control}. The leg length actuation replicates the knee flexion and extension during the swing phase, while during the double-support it leads to a stabilizing energy cycling \cite{rezazadeh2020control}. In this study, we consider three variations of models based on this paradigm in the frontal plane: (i) fixed hip, (ii) fixed ankle, and (iii) both ankle and hip free to move. Fig.~\ref{SLIP_model} shows detailed schematics for these model variations.

\begin{figure}
	\centering
	\includegraphics[width=0.9\linewidth]{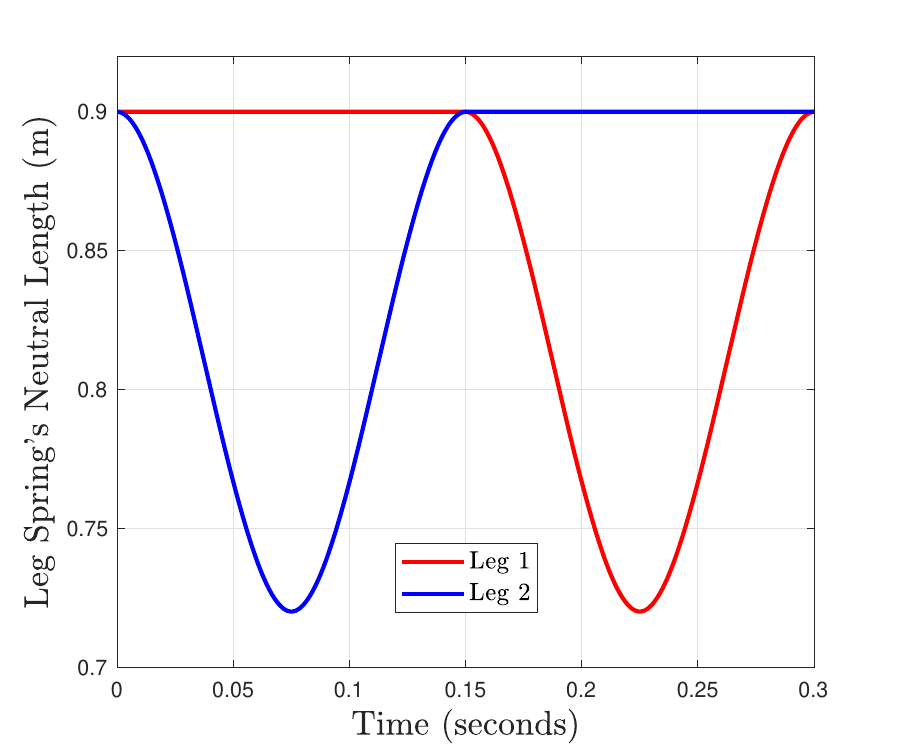}
	\caption{Profile of the neutral spring length for a stride time of 0.3 seconds and the maximum (stance phase) resting length of $l_0=0.9$ m. The maximum leg retraction is 0.18 m (1/5 of the maximum resting leg length).}
	\label{spring_length_vel_profile}
\end{figure}

The dynamics of the models are primarily governed by the forced oscillations generated by the periodic changes of the neutral length of the spring-damper system:
\begin{equation}
	F=k(l_n-l)+b(\dot{l}_n-\dot{l}),
\end{equation}
\noindent where $F$ is the spring-damper force, $l$ is the leg length, and $l_n(t)$ is the neutral length of the spring, which changes periodically according to Fig.~\ref{spring_length_vel_profile}. The frequency can be increased or decreased by adjusting the leg length actuation rate, which is determined by the stride time of the model. In \cite{rezazadeh2015spring}, a sinusoidal profile was suggested for the swing retraction-extension part, and a constant value $l_n=l_0$ roughly corresponding to the stance part. In this work, we changed the sinusoidal profile to a quintic spline to bring the leg velocity closer to zero at foot contact. As a result of this, the damping force is reduced at touchdown, producing smoother transitions between the single and double support phases. We first consider two simplified models (fixed-hip and fixed ankle) which provide insights in the stability mechanisms and the effect of parameters, and then compare the results with a more general case.

\subsubsection{Fixed-Hip Model}
In the fixed-hip case (Fig. \ref{SLIP_model}(a)), the hip motion is constrained, and the legs always remain perpendicular to the hip. During single support, the stance foot acts as a pivot point and provides two degrees of freedom (DoF). In double support, however, both legs are fixed to the ground, reducing the system to one DoF, where the CoM can only move along the leg length direction, as the legs are constrained to be parallel due to the fixed hips.

\subsubsection{Fixed-Ankle Model}
In this configuration (Fig. \ref{SLIP_model}(b)), the CoM is restricted not to rotate around the stance foot’s ankle joint. During single support, the system has two DoF due to hip joint motion. As in the fixed-hip case, the DoF reduce to one during double support because both legs are constrained to the ground. Additionally, the swing leg is assumed to remain perpendicular to the ground, and no hip torque is required to place it in the desired position, since the legs are modeled as massless.

\subsubsection{Free Ankle and Hip Model}
In the final configuration (Fig. \ref{SLIP_model}(c)), both the hip and the ankle are free to move. Inspired by mass distribution between upper and lower body in humans which was replicated in the Mithra humanoid robot \cite{winter2009biomechanics, semasinghe2025design}, we assign 30\% of the total body mass to the legs, with the torso carrying the remaining 70\%. The leg mass is positioned below the spring-damper elements at a fixed offset from the foot. Likewise, the torso inertia was considered and applied using an anthropometric value. The model possesses 7 DoF in flight, 5 DoF in single support, and 3 DoF in double support. The reduction in DoF across phases arises from the constraints imposed by foot–ground contact.

In this study, we assume that: (i) the model is constrained to move only in the frontal plane; (ii) the stance foot remains fixed to the ground without slipping until the transition to the swing phase; and (iii) the legs are considered massless in the fixed-ankle and fixed-hip cases. The first assumption is supported by empirical studies on human locomotion \cite{bauby2000active,perry2017walking}, which show that step length does not change significantly with step width variation, indicating minimal coupling between sagittal and frontal plane dynamics.

\subsection{Model Dynamics}
The dynamics of the frontal-plane models are derived using the Euler-Lagrange method. 
The assignment of the generalized coordinates across the three models is specified as follows: $q = [l_{st} \hspace{5pt} \theta_{a_{st}}]^T$ for the model with fixed hips, $q = [l_{st} \hspace{5pt} \theta_{h_{st}}]^T$ for the fixed ankles, and $q = [x \hspace{5pt} y \hspace{5pt} l_{st} \hspace{5pt} \theta_{a_{st}} \hspace{5pt} \theta_{h_{st}} \hspace{5pt} l_{sw} \hspace{5pt} \theta_{h_{sw}}]^T$ is used for the model in which both the hip and ankle are allowed unrestricted movement. In this framework, $x$ and $y$ represent the horizontal and vertical positions of the torso mass, while subscripts $st$ and $sw$ are used to differentiate between the stance leg and the swing leg, respectively. The definitions of the terms encompassed in the generalized coordinates are illustrated in Fig.~\ref{SLIP_model}.

\subsection{Stability of the Periodic Orbits}
We implemented the models in MATLAB R2024b and solved using \texttt{ode45}. 
To identify the periodic orbits, we used the \texttt{fminsearch}. 
The stability of the computed periodic orbit was evaluated using the eigenvalues of the linearized Poincaré map after introducing small perturbations to the Poincaré section states. If the absolute value of the maximum eigenvalue is less than 1, the Poincaré section of the model and thus the periodic orbit are considered stable. The minimum stabilizing stride frequency was obtained by identifying the minimum frequency of the neutral spring length (Fig. \ref{spring_length_vel_profile}) at which a symmetric, stable cycle with no flight phase is emerged.

\section{Results}

In this section, we present the simulation results for various parameter combinations and their effects on the minimum required  stride frequency for stability. The focus of the analyses is on how the forced-oscillation scheme for the legs can generate stable periodic motions. For consistency, the values of the following parameters were kept fixed in all cases: a leg damping ratio of 0.1, a maximum spring compression equal to $1/5$ of the maximum resting length, and a torso mass offset of 0.2 m.

\subsection{Effect of Mass and Inertia}
\begin{figure}
	\centering
	\subfigure[]{\includegraphics[width=0.9\linewidth]{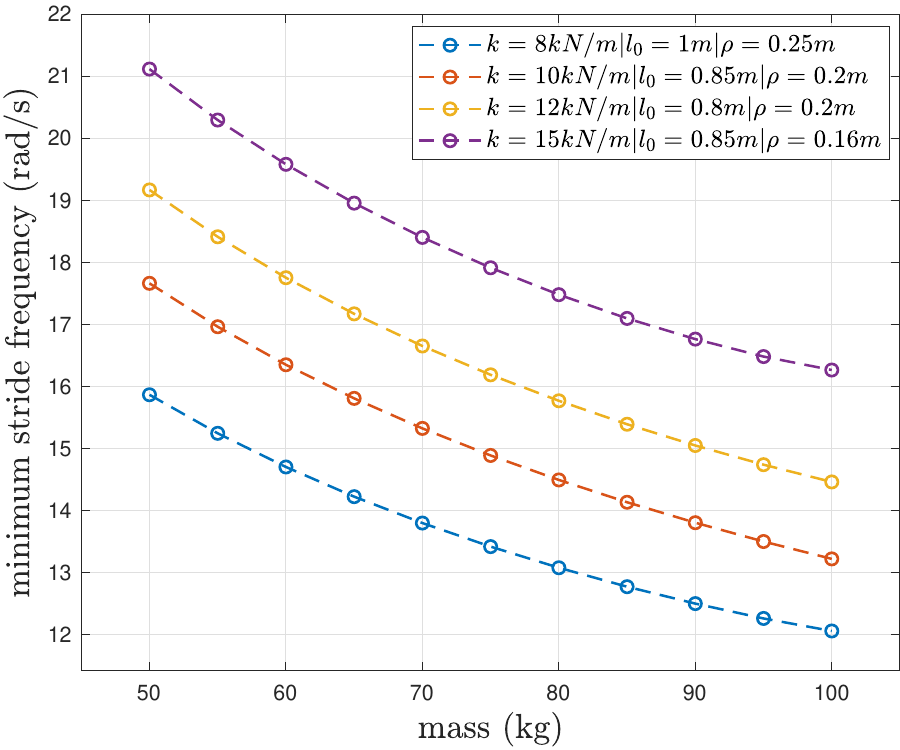}\label{m_plot_fixed_hip}}
	\subfigure[]{	\includegraphics[width=0.9\linewidth]{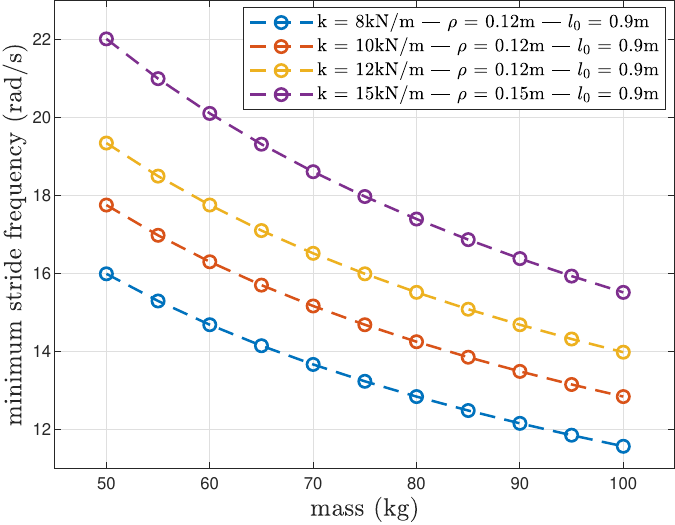}\label{m_plot_fixed_ankle}}
	\caption{Effect of mass on the minimum stride frequency for stability; (a) fixed hip, and (b) fixed ankle configurations. The mass of each model varies from 50 kg to 100 kg.}
	\label{m_plots}
\end{figure}

We simulated the system for different sets of mass, leg stiffness, maximum spring rest length, and hip width for both fixed-hip and fixed-ankle cases, and determined the minimum frequency that leads to stable oscillations. In Fig.~\ref{m_plots}, it can be observed that as the mass increases, the minimum stride frequency required for stability decreases, regardless of the model configuration. The plots reveal a nonlinear relationship between the mass and the minimum stride frequency, with a consistent trend across all plots.


\begin{figure}
	\centering
	\includegraphics[width=0.93\linewidth]{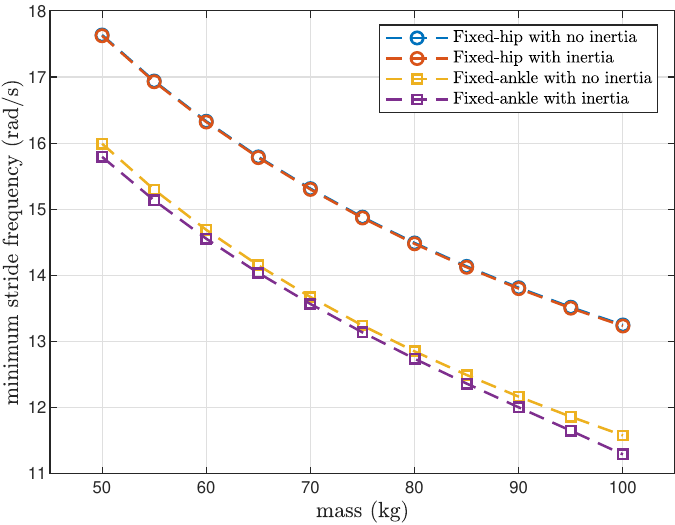}
	\caption{Variation of the minimum stride frequency without and with torso inertia for the fixed-hip and fixed-ankle models.}
	\label{inertia_plot}
\end{figure}

In addition, we analyzed the effect of inertia by introducing a torso roll inertia. A constant radius of gyration of 0.3 m was assumed in all models and the variation in the minimum stride frequency was compared between models with and without inertia. As shown in Fig.~\ref{inertia_plot}, the inclusion of inertia does not significantly affect the stability, as the minimum stride frequencies remain nearly identical.

\subsection{Effect of Leg Stiffness}
\begin{figure*}
	\centering
	\subfigure[]{\includegraphics[width=0.45\linewidth]{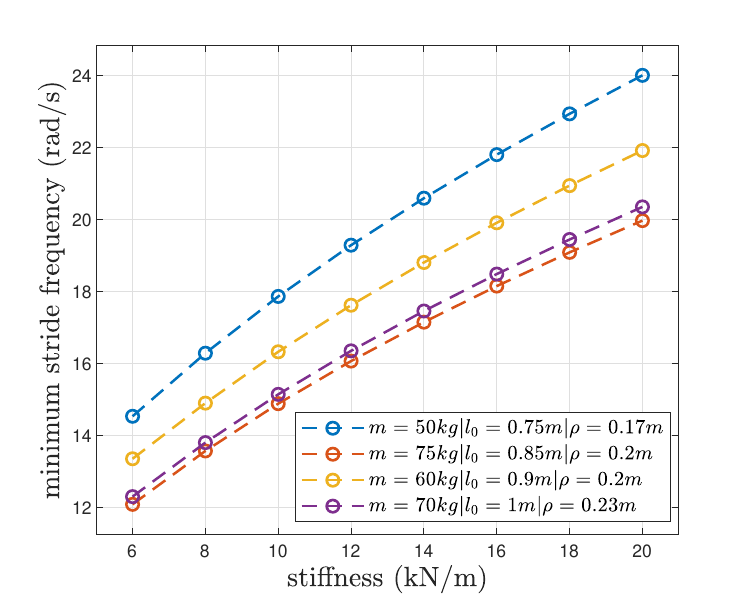}\label{k_plot_fixed_hip}}
	\subfigure[]{	\includegraphics[width=0.45\linewidth]{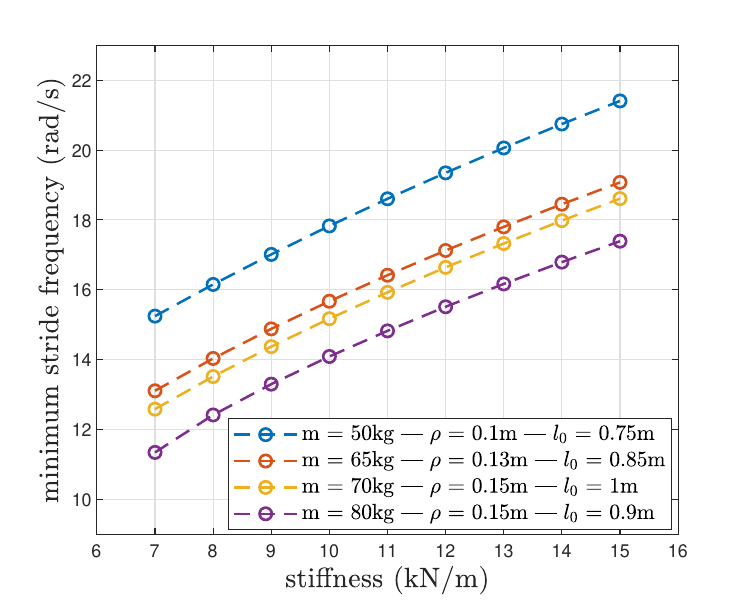}\label{k_plot_fixed_ankle}}
	\caption{Effect of leg stiffness on the minimum stabilizing stride frequency; (a) for hip-fixed and (b) for ankle-fixed configurations.}
	\label{k_plots}
\end{figure*}

We analyze the effect of leg stiffness on the minimum stride frequency in both the hip-fixed and ankle-fixed models. In each simulation, the stiffness of the spring was varied from 6 kN/m to 20 kN/m, as this range include the values observed in human walking \cite{geyer2006compliant}. As shown in Fig.~\ref{k_plots}, increasing leg stiffness leads to a greater minimum stride frequency compared to lower stiffness values. Doubling stiffness increases the minimum frequency by approximately 30–40\% in all models. This highlights that leg stiffness has a significant effect on the frontal-plane stability, as stiffer-legged models require quick stepping to maintain stable locomotion compared to models with lower stiffness.


\subsection{Effect of Hip Width}
\begin{figure*}
	\centering
	\subfigure[]{\includegraphics[width=0.45\linewidth]{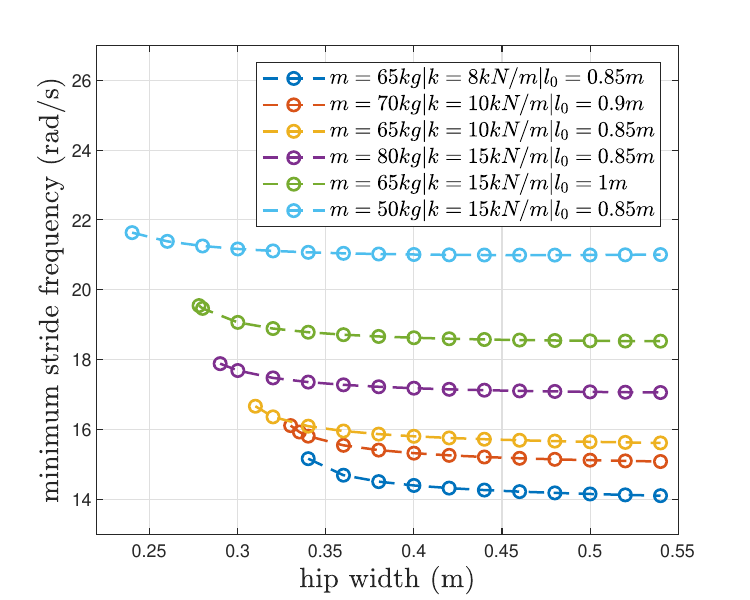}\label{w_plot_fixed_hip}}
	\subfigure[]{	\includegraphics[width=0.45\linewidth]{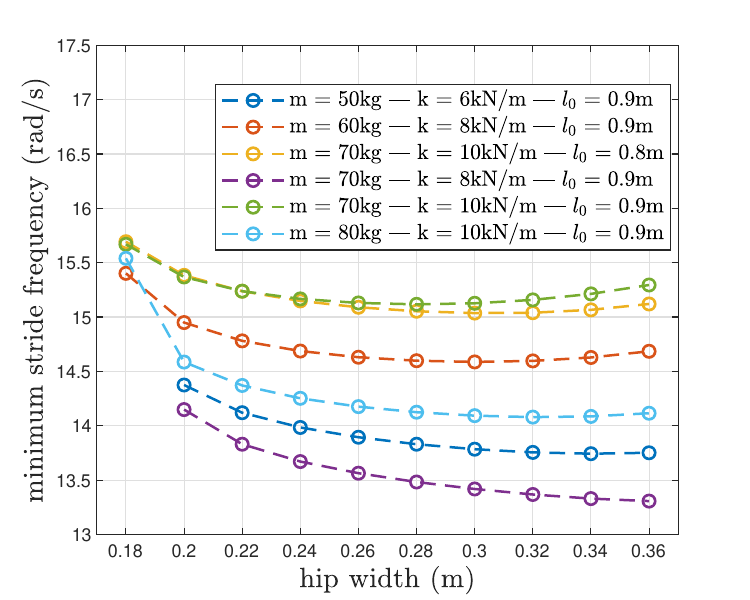}\label{w_plot_fixed_ankle}}
	\subfigure[]{\includegraphics[width=0.45\linewidth]{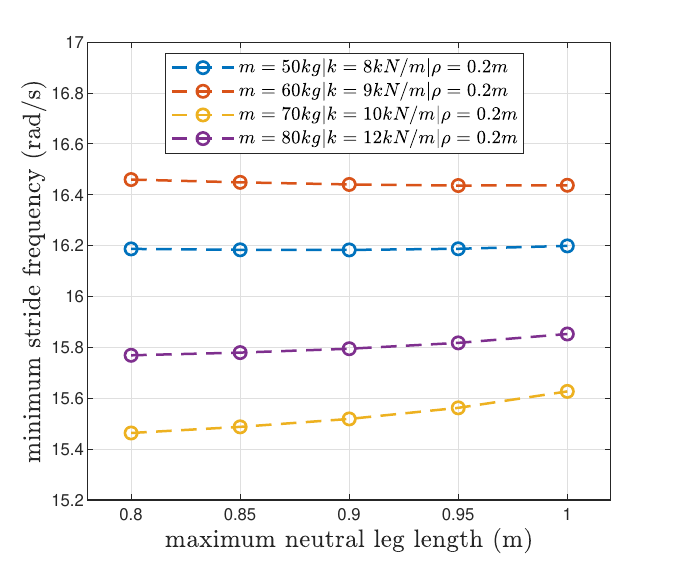}\label{l_plot_fixed_hip}}
	\subfigure[]{	\includegraphics[width=0.45\linewidth]{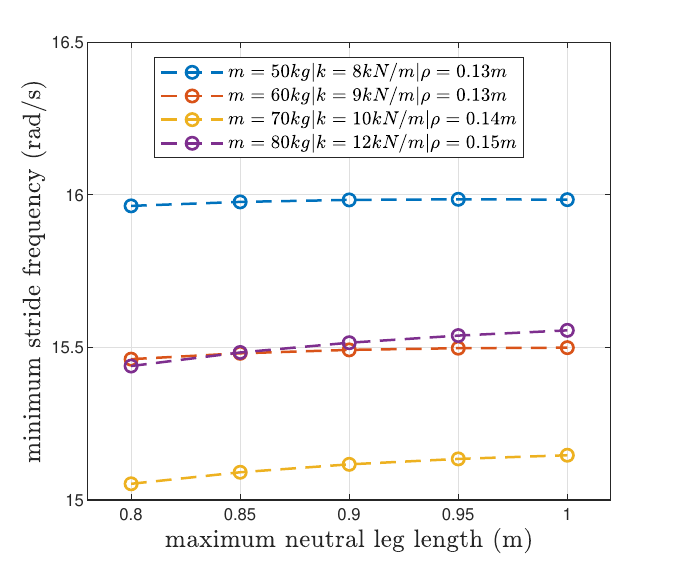}\label{l_plot_fixed_ankle}}
	\caption{Variations of minimum stride frequencies with model length parameters: (a) hip width, fixed-hip model; (b) hip width, fixed-ankle model; (c) stance resting leg length $l_0$, fixed-hip model; (d) stance resting leg length $l_0$, fixed-ankle model.}
	\label{w_plots}
\end{figure*}

Hip width is another parameter that can influence the frontal-plane stability. This offset induces passive motion around the ankle and hip joints of the stance leg during single support, allowing the mass to fall under gravity until the swing leg makes ground contact. Consequently, the CoM oscillates back and forth around the vertical axis, producing a periodic pattern observed in every simulation. For these simulations, we focused on hip widths ranging from 0.18 m to 0.54 m. As shown in Fig.~\ref{w_plots}, reducing the hip width increases the minimum stride frequency, with the effect being more pronounced at smaller widths. For example, in the fixed-hip model with a natural frequency of $11\text{ rad/s}$, reducing the hip width from $0.36\text{ m}$ to $0.34\text{ m}$ increases the minimum stride frequency by 3\%, whereas a similar reduction from $0.54\text{ m}$ to $0.52\text{ m}$ increases it by only 0.1\%. In comparison, a model with a natural frequency of $14\text{ rad/s}$ shows a 1\% increase for the same narrow-width reduction and only a 0.04\% increase for the wider-width reduction. This demonstrates that the sensitivity of the minimum stride frequency to hip width is greatest in narrower configurations, particularly for models with lower natural frequencies. As the model’s natural frequency increases, this sensitivity diminishes, and the minimum stride frequency converges toward a nearly fixed value across all hip widths.

In contrast, in the fixed-ankle model, both lower and higher natural frequency cases show great sensitivity at narrow hip widths (Fig.~\ref{w_plot_fixed_ankle}). However, as the natural frequency increases, the required stepping frequency at wider hip widths must also increase to maintain stability, unlike models with lower natural frequencies. As such, the results indicate that variations in hip width have a weaker influence on reducing the minimum stride frequency compared to the changes in mass or leg stiffness.


\begin{figure}
	\centering
	\includegraphics[width=1\linewidth]{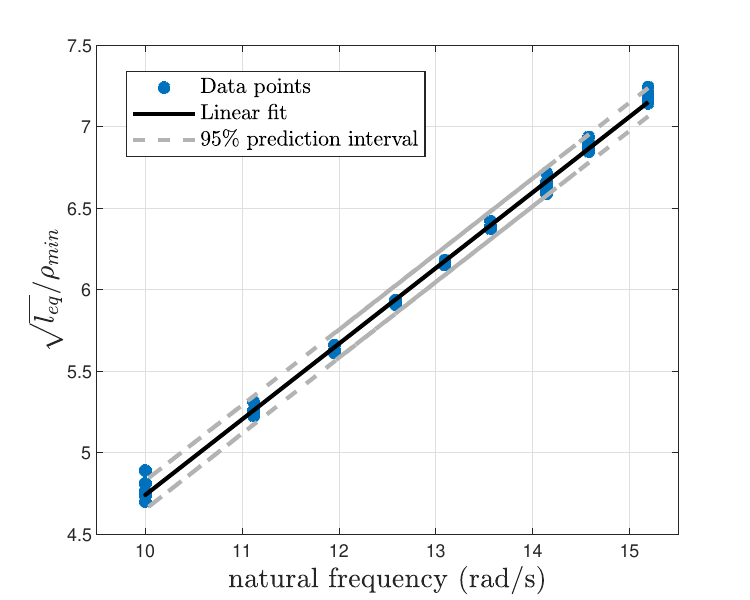}
	\caption{Linear relationship between natural frequency and the minimum hip width for the existence of a stable periodic orbit in the fixed-hip models.}
	\label{min_step_width_plot}
\end{figure}

We also found that as the hip width approaches zero, a threshold value appears below which the model’s periodic orbits become unstable, while widths above this threshold yield stable orbits. This behavior is observed in both fixed-hip and fixed-ankle models. In particular, the threshold value for the fixed-hip case is greater than that for the fixed-ankle case. Moreover, in the case of fixed hips, we were able to establish a relationship between this threshold value and other model parameters, while no simple relationship could be identified for the case of fixed ankles. 
To find this relationship, we simulated models with different natural frequencies using five different resting leg lengths, ranging from $0.7\text{ m}$ to $1.2\text{ m}$, and for each case, the minimum hip width was identified for the respective model. As shown in Fig.~\ref{min_step_width_plot}, the data points reveal a linear relationship between the natural frequency of the model $\omega_n$ and $\sqrt{l_{eq}}/\rho_{min}$. Here, $l_{eq}$ is defined as $l_{eq} = l_0 - \delta l$, where $l_0$ is as in Section 2.1, $\rho$ is half of the width of the hip and $\delta l = mg/k$. Using $l_{eq}$ instead of $l_0$ provides a more general formulation, as it incorporates the effect of mass and gravity on the effective resting leg length. The linear fit yields $R^2 = 0.997$, with a slope of approximately 0.5 and an intercept near zero, leading to the following expression.  
\begin{equation}
	\frac{\sqrt{l_{eq}}}{\rho_{min}}  \approx  \frac{\omega_n}{2}, 
	\label{eq_approx}
\end{equation}
\noindent where $\omega_n = \sqrt{\frac{k}{m}}$. Since the minimum hip width $w_{min}$ can be defined as $2\rho_{min}$, the final formulation yields Eqn.~\eqref{eq_min_rho}.
\begin{equation}
	w_{min} \approx 4\frac{\sqrt{l_{eq}}}{\omega_n}. 
	\label{eq_min_rho}
\end{equation}

\subsection{Effect of Maximum Resting Leg Length}
In all three cases, the spring resting leg length $l_n$ is a time-dependent variable, as shown in Fig.~\ref{spring_length_vel_profile}. This in turn affects the real leg length during stance phase, which is determined by the deflection of the spring. Given that the hip width is small compared to the leg length, the latter has a significant influence on the frequency exhibited by the model during the single support phase, as the system functions as an inverted pendulum anchored at the stance foot. Figures~\ref{l_plot_fixed_ankle} and ~\ref{l_plot_fixed_hip} demonstrate that in both fixed-ankle and fixed-hip scenarios, changes in the maximum (stance) resting leg length $l_0$ does not substantially impact the minimum stride frequency, as evidenced by the minimal variation observed between the data points. Furthermore, the analysis reveals that, across several plots, an increase in leg length is associated with a small increase in the minimum stride frequency, implying that longer leg lengths may be less stable. This phenomenon could potentially be ascribed to the hip width incorporated in the model; an increase in hip width may facilitate the discovery of more stable configurations for models characterized by extended leg lengths.


\subsection{Effect of Natural Frequency}
We randomly generated 50 models with natural frequencies ranging from 8 rad/s to 20 rad/s. This range was selected to ensure that the corresponding stiffness and mass values fall within the desired range for the analysis. For each natural frequency, stiffness was randomly selected and the corresponding mass was computed accordingly. The maximum resting leg length was also randomly chosen, and the minimum hip width required for stability was calculated using Eqn.~\eqref{eq_min_rho}. To ensure the existence of a stable periodic orbit, the final hip width for each model was set to be greater than this minimum value. The ranges of the parameters used to generate the models are summarized in Table~\ref{parameter_table}.

\begin{table}[h]
	\caption{\label{parameter_table}Ranges of parameters used to generate the models for the analysis of the effects of the natural frequency. The hip width is not assigned a fixed range, as it is selected individually for each model to be greater than the minimum required for stability.}
	\centering{%
		\begin{tabular}{l c c}
			\toprule
			Parameter & Minimum value & Maximum value  \\ 
			\midrule
			Mass ($m$) & 50 kg & 100 kg  \\
			Stiffness ($k$) & 6 kN/m & 20 kN/m \\
			Resting leg length ($l_0$) & 0.7 m & 1.2 m  \\
			\bottomrule
		\end{tabular}
	}%
\end{table}

\begin{figure*}
	\centering
	\subfigure[]{\includegraphics[width=0.45\linewidth]{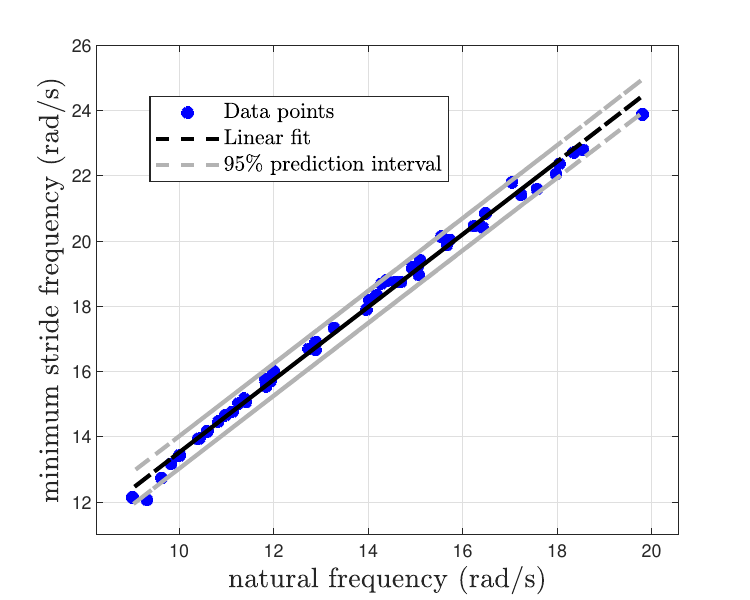}\label{wn_plot}}
	\subfigure[]{	\includegraphics[width=0.45\linewidth]{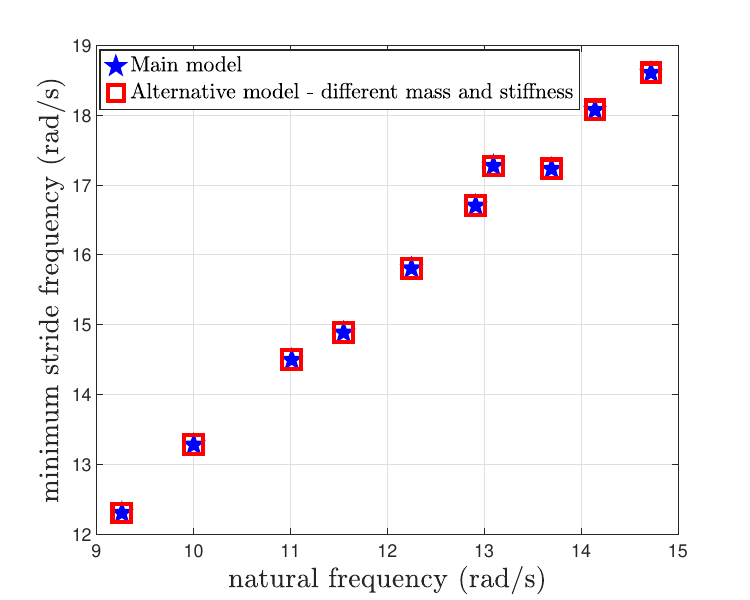}\label{same_wn_plot}}
	\caption{Effect of hip width on the minimum stabilizing stride frequency; (a) for fixed-hip, and (b) for fixed-ankle configurations. The minimum value of each plot shows the threshold value of step width for the existence of a stable periodic pattern.}
	\label{wn_plots}
\end{figure*}

Figure~\ref{wn_plot} illustrates the distribution of data points derived from 50 randomly generated models, which elucidates the relationship between the model's natural frequency and the minimum stride frequency. The use of linear regression on these data points demonstrates that there is a linear correlation between the increase in the natural frequency of the model and the increase in the minimum stride frequency, as indicated by a linear fit $R^2$ value of 0.99. Additionally, the narrow width of the 95\% prediction interval indicates that the model exhibits precision and low data variability.


Figure~\ref{same_wn_plot} illustrates that maintaining the same natural frequency, regardless of different combinations of mass and stiffness, results in an identical minimum stride frequency for a model with the same resting leg length and hip width. This observation suggests that the models' natural frequencies are unique determinants of their stability margins. Consequently, it becomes feasible to design equivalent models that achieve comparable stability, even when constraints are imposed on mass and stiffness selection.

\subsection{Minimum Stride Frequency Prediction for Feedforward Stability}
Motivated by the relation between the model's natural frequency and its minimum stabilizing stride frequency, it was posited that the latter could be predicted through the utilization of model parameters. Consequently, two dimensionless parameters were defined for comparative analysis between models: dimensionless leg stiffness ($\hat{k}$) and dimensionless stride frequency ($\hat{\omega}_s$). These parameters are defined as follows: $\hat{k} = {kl}/{mg}$, where $l = \sqrt{l_0^2 + \rho^2}$, and $\hat{\omega}_s = \omega_s /\sqrt{{g}/{l}}$, where $\omega_s$ represents the stride frequency and $g$ denotes gravity.

For the generation of data points pertinent to this analysis, a consistent cohort comprising 50 randomly generated models was utilized. As illustrated in Fig.~\ref{dimensionless_plots}, the dimensionless stride frequency \(\hat{\omega}_s\) demonstrates a linear correlation with $\sqrt{\hat{k}}$, with an $R^2 = 0.99$. This relationship is characterized by a slope which approximately equal to 1 and an intercept near 1. Consequently, this relationship can be succinctly approximated as:
\begin{equation}
	\hat{\omega}_s \approx \sqrt{\hat{k}} + 1.
	\label{eq_stride_prediction}
\end{equation}
\noindent Upon substitution of the dimensionless terms with the respective model parameters, Eqn. \eqref{eq_stride_prediction} yields;
\begin{equation}
	\omega_{s,min} \approx \omega_n + \omega_p.
	\label{eq_min_stride_f}
\end{equation}
\noindent Here, $\omega_p$ is defined as $\sqrt{g/l}$, which signifies the frequency of an inverted pendulum in single support. Consequently, Eqn. \eqref{eq_min_stride_f} indicates that the minimum stride frequency can be precisely approximated by considering the sum of the model's natural frequency due to leg compliance and the pendulum frequency of the model during its single support phase. Moreover, the data points are in close alignment with the linear fit line, and the narrow 95\% prediction interval increases the probability of the predictions falling within this band. Therefore, it is suggested that through this relationship (i.e., Eqn.~\eqref{eq_min_stride_f}), the minimum stride frequency for any specified model can be computed, which facilitates the selection of a stride frequency that ensures the model stability.



\subsection{Comparison with the Extended Model}
In previous subsections, we evaluated the stability of the fixed-ankle and fixed-hip models through their minimum stabilizing stride frequencies. These models limit the motions of the hip or ankle to simplify the dynamics of the model, and at the same time, provide insight for the stability of full-order systems. To better capture realistic dynamics in the frontal plane, we used a variation of the model closer to full-order systems, as depicted in Fig.~\ref{SLIP_model}(c). 

\begin{figure}
	\centering
	\includegraphics[width=.9\linewidth]{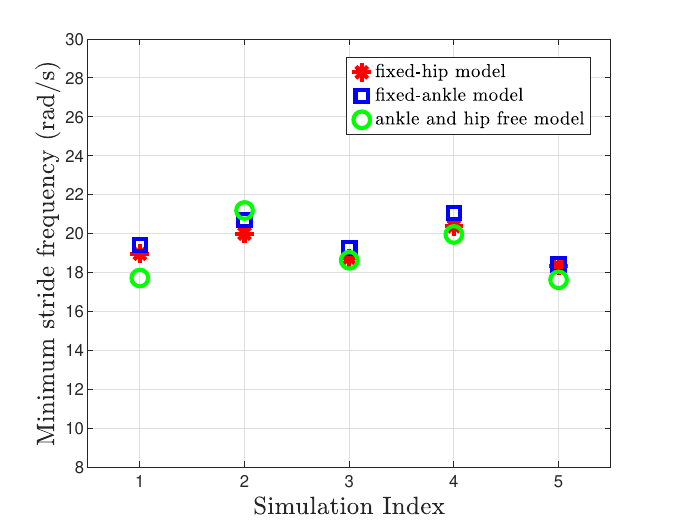}
	\caption{Comparison of lateral stability in the three models. The minimum stabilizing frequencies are close to each other in all three cases.}
	\label{model_compare_passive}
\end{figure}

As presented in Fig.~\ref{model_compare_passive}, the evaluation of the stability was conducted for all three configurations for five different sets of model parameters. The fixed-ankle and fixed-hip configurations exhibited nearly identical minimum stride frequencies, with a variation of less than 1\%. It is noteworthy that the minimum stride frequency of the extended model is closely aligned with that of the simpler configurations. In the simulation of the extended model, no active control was exerted on the stance leg's ankle and hip joints. However, given that the swing leg in this configuration possesses mass, in contrast to the previous two models, a basic Proportional-Derivative (PD) was introduced to the swing leg's hip to maintain the swing foot's perpendicularity to the ground. The controller gains included a proportional gain $K_P$ selected within the specified range $300-500\text{ Nm/rad}$ (from the human hip quasi-stiffness values, as reported in \cite{shamaei2013estimationhip, molitor2024lower}), while the derivative gain $K_D$ was calculated using the damping ratio of $0.1$.


\subsection{Combination with Active Stance Phase Control}
Building upon our prior investigation of feedforward stability in the three proposed models, we incorporated active control into each model to assess whether their stability could be improved through introducing a joint level PD controller for each moving joint. In the fixed hip model, the ankle joint of the supporting leg was controlled to remain vertical, while in the fixed ankle model, the hip joint was controlled to keep the pelvis parallel to the ground. To represent the human quasi-stiffness at the ankle joint~\cite{shamaei2013estimation, molitor2024lower}, $K_P$ of the PD controller in the range of $100$-$200 \text{ Nm/rad}$ was selected. Likewise, the proportional gain for the hip joint controller was chosen in the human joint quasi-stiffness range, as previously discussed. For both controllers, the derivative gain was determined by using a damping ratio of 0.1. Given that the legs are massless in the two simplified models, no actuation is applied to the swing leg.

\begin{figure}
	\centering
	\includegraphics[width=.9\linewidth]{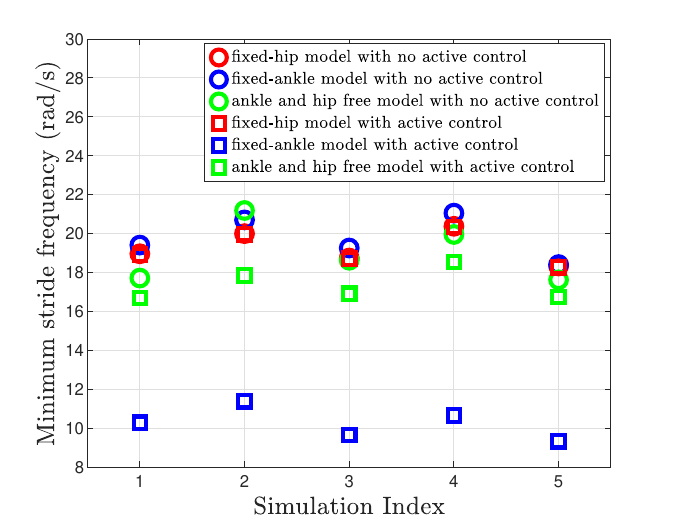}
	\caption{Imrpovement of lateral stability in the three models with active stance phase control.}
	\label{model_compare_active}
\end{figure}

As shown in Fig.~\ref{model_compare_active}, the fixed-hip model shows no significant improvement (less than 1\%) even with active ankle control. In contrast, the fixed-ankle model demonstrates a substantial improvement, with the minimum stride frequency for stability reduced by approximately 85\% in all cases. To maintain consistency between simulations, we used $K_P = 150\,\text{Nm/rad}$ for the ankle joint controller in the fixed-hip model, and $K_P = 300\,\text{Nm/rad}$ for the hip joint controller in the fixed-ankle model. Similarly, for the extended model (with both ankle and hip free to move), we used the same gains as before, except for the swing leg. The gain $K_P$ of the hip controller for the swing leg was increased to $500\text{ Nm/rad}$ to achieve accurate tracking of the desired foot placement, ensuring that the swing leg remained vertical during the swing phase. The model's stability was improved by approximately 15\% relative to the active control of the fixed-hip case. However, the same minimum stride frequency achieved by the ankle-fixed model remained unattainable, since increasing the controller gains led to instability.


\section{ Discussion}

Lateral stability is significantly influenced by the frequency of the legs and the model's mass and leg stiffness. A greater body mass provides better lateral stability than a smaller body mass. In human locomotion, leg stiffness generally increases with an increase in speed \cite{arampatzis1999effect}. This observation is also relevant in the current context, in which an increase in leg stiffness is associated with a higher minimum stride frequency, indicating the need for faster stepping to maintain stability. Given that the body mass is a constant parameter in humans, adaptations in leg stiffness in correlation with movement speed may contribute to improved stability in the frontal plane.

The stance neutral leg length and the hip width, conversely, make smaller contributions to the lateral stability. The simulation data reveal that a minimum hip width is required to achieve a stable periodic orbit. This aligns with the fact that, in the sagittal plane (where hip width is zero), maintaining stability through feedforward schemes alone is not possible, and requires stance-phase and/or foot-placement control. On the other hand, leg length and hip width influence the additional frequency attributes of the models, capturing the inverted pendulum characteristics of the models during the single-support phase.

It is important to mention that our findings are closely aligned with several empirical studies on human locomotion. The study presented in \cite{perry2017walking} indicates that an increase in step width does not significantly enhance frontal plane stability. Although step width was not considered in our study, we observed that an increase in hip width does not lower the minimum stride frequency of the model in either the fixed-hip or fixed-ankle configurations (Figs.~\ref{w_plot_fixed_hip} and ~\ref{w_plot_fixed_ankle}). Furthermore, studies \cite{donelan2004mechanical} and \cite{ijmker2013energy} indicate that external stabilization in the pelvis enables humans to adopt a narrower step width, as this stabilization mitigates the need for an active controller. In our fixed-ankle model, a similar approach was used by restricting the rotation about the ankle joint and thereby restricting pelvic motion. As observed, smaller hip widths in this case can yield similar stability results to those of the fixed-hip model with larger hip widths.

We found that the natural frequency significantly influences the lateral stability. Additionally, the impact of mass and leg stiffness were separately demonstrated through their individual effects, as evidenced by the significant variations observed in the minimum stride frequency. Moreover, it is highlighted that given equivalent leg lengths and hip widths, any combination of mass and stiffness that results in the same natural frequency will yield similar outcomes. This result is crucial in the design of bipedal models where there are constraints on the selection of mass and stiffness, as alternative configurations produce comparable results.

\begin{figure*}
	\centering
	\subfigure[]{\includegraphics[width=0.45\linewidth]{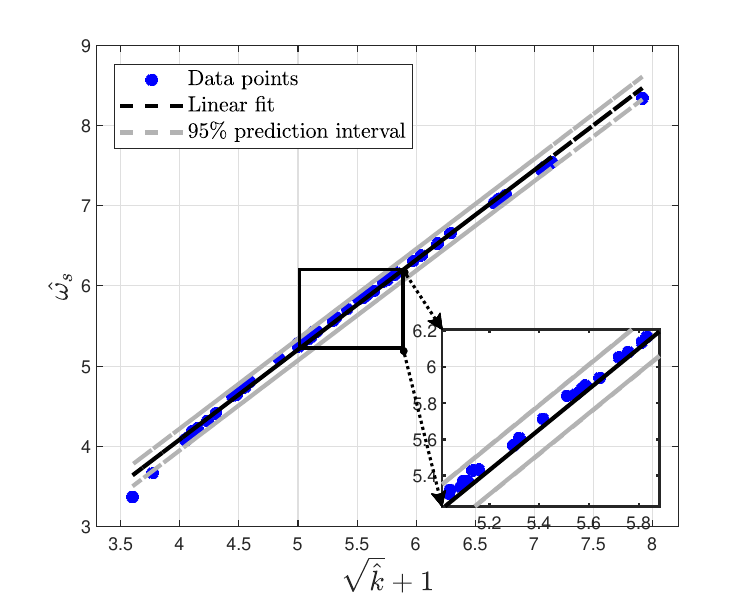}\label{dimensionless_plots}}
	\subfigure[]{	\includegraphics[width=0.45\linewidth]{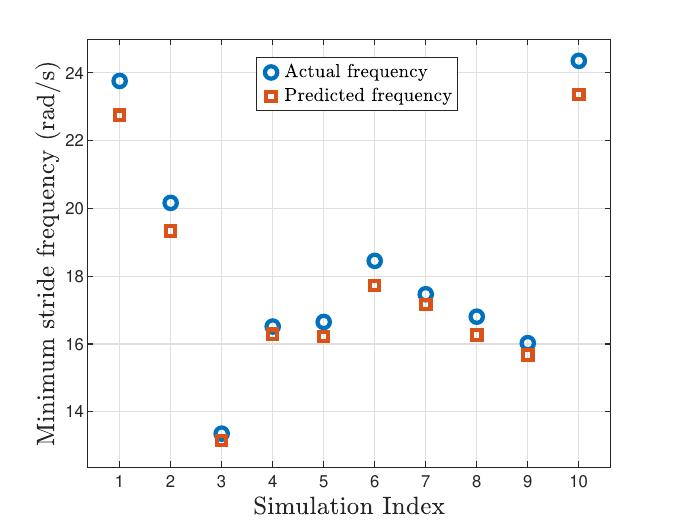}\label{prediction_plots}}
	\caption{(a) Linear corelation between the dimensionless stiffness $\hat{k}$ and the dimensionless stride frequency $\hat{\omega_s}$ for 50 sets of model parameters using the Monte Carlo method. (b) Validation of the proposed hypothesis of predicting minimum stride frequency based on 10 different sets of model parameters.}
	\label{final_prediction_plots}
\end{figure*}

In this study, our primary objective was to establish a quantitative framework for assessing frontal-plane stability in relation to model parameters. The results from 50 simulations with different sets of parameters indicate that the minimum stride frequency can be approximated as the sum of the model’s natural frequency and an additional pendulum frequency component. The natural frequency emerges as the dominant factor in this relationship, since the pendulum frequency during single support is comparatively small. Figure~\ref{prediction_plots} further illustrates this by comparing predicted and actual minimum stride frequencies for 10 randomly generated models. The differences are minimal, with the largest deviation being about 1 rad/s (0.16 Hz), thereby confirming both the accuracy and reliability of the proposed formulation.

In order to assess these results in a case more closely matching the dynamics of a full-order system, we extended our analyses to a more generalized 7-DoF model. The 7-DoF model includes leg mass, allowing it to more closely replicate human dynamics in the frontal plane compared to the simpler models. Using the same set of model parameters, all three models produced very similar minimum stride frequencies, highlighting that the feedforward stability is preserved. This indicates that results from the proposed simple model can reliably be used to assess the stability of the more complex model, as their outcomes are closely aligned. Furthermore, it demonstrates that feedforward stability depends primarily on the model parameters rather than the specific configuration, such as the foot placement strategy, since the foot placements among the models are not identical.


Empirical studies on human walking indicate that lateral balance requires active control to maintain stability. To capture this, we implemented simple PD controllers on each moving joint of the models, keeping the gains small and within the range of human joint quasi-stiffness. In the fixed-hip model, active control had little effect, as the small proportional gain produced insufficient torque to maintain the stance leg vertical. In contrast, the fixed-ankle model showed a substantial improvement in stability with only small hip control, likely because fixing the ankle restricts pelvic motion and ensures the swing leg contacts the ground vertically, reducing step width variations. As expected, for the extended model (both ankle and hip free to move), stability improved more than the hip-fixed case, but due to pelvic motion and foot placement variations did not reach the level of the fixed-ankle model. We also observed that increasing controller gains beyond this small range often led to instability in all models, indicating that overly aggressive PD control can destabilize the system. Nonetheless, the feedforward stability schemes discussed can substantially reduce control effort and enhance robustness against external perturbations by relying less on feedback. 

\vspace{3mm}
\textit{Limitations}
\vspace{1mm}

We presented our results based on the frontal-plane dynamics of SLIP-based models in in-place stepping. In our study, we assumed that sagittal-plane motion is restricted and that frontal-plane can be decoupled from the sagittal-plane dynamics. However, previous studies have shown that a slight coupling exists between sagittal- and frontal-plane locomotion \cite{donelan2004mechanical,ijmker2013energy}. Therefore, this study may be extended to include coupled sagittal-plane dynamics and investigate how they affect lateral stability. Additionally, we found that simple PD controllers with fixed gains are not sufficient for effective active control, and more sophisticated control strategies are needed. Since the quasi-stiffness values of human joints change during the gait cycle \cite{shamaei2013estimation, shamaei2013estimationhip}, including variable stiffness profiles can potentially lead to improving the stability characteristics.

\section{Conclusion}
This study demonstrated how frontal-plane stability was influenced by model parameters such as mass, leg stiffness, hip width, and leg length. We qualitatively evaluated and presented the stability in terms of the minimum stride frequency at the boundary between stability and instability of the models. Our results show that when using a periodic leg retraction and extension scheme, in models with different levels of complexity and without any feedback control, the frontal-plane dynamics exhibit stable periodic orbits. Based on these findings, we proposed a formula to predict the minimum stride frequency for feedforward stability using model parameters. We showed that the results from simple models in fact closely align with those of the extended model, which can facilitate the analysis of models with higher complexities using simple models. The introduction of active controllers further improved the lateral stability and led to requiring lower feedforward frequencies. As such, the feedforward scheme and the feedback controllers can be designed together for optimal robustness and effort. Based on these points, the presented approach provides a framework for better understanding of lateral stability in bipedal locomotion and more efficient control design without the need for extensive analyses of complex systems.


\nocite{*} 

\bibliographystyle{asmejour}   

\bibliography{journal_ref} 

\begin{thebibliography}{10}
\newcommand{\enquote}[1]{``#1''}
\providecommand{\url}[1]{\texttt{#1}}
\providecommand{\urlprefix}{}
\expandafter\ifx\csname urlstyle\endcsname\relax
  \providecommand{\doi}[1]{doi:\discretionary{}{}{}#1}\else
  \providecommand{\doi}{doi:\discretionary{}{}{}\begingroup
  \urlstyle{rm}\Url}\fi
\providecommand{\eprint}[2][]{\urlprefix\url{#1#2}}
\providecommand{\hrefurl}[2]{\href{#1}{#2}}

\bibitem{mille2005age}
Mille, M.-L., Johnson, M.~E., Martinez, K.~M., and Rogers, M.~W., 2005,
  \enquote{Age-dependent differences in lateral balance recovery through
  protective stepping,} Clinical Biomechanics, \textbf{20}(6), pp. 607--616.

\bibitem{kajita2019position}
Kajita, S., Benallegue, M., Cisneros, R., Sakaguchi, T., Morisawa, M.,
  Kaminaga, H., Kumagai, I., Kaneko, K., and Kanehiro, F., 2019,
  \enquote{Position-based lateral balance control for knee-stretched biped
  robot,} \textit{2019 IEEE-RAS 19th International Conference on Humanoid
  Robots (Humanoids)}, pp. 17--24.

\bibitem{ito2008zmp}
Ito, S., Amano, S., Sasaki, M., and Kulvanit, P., 2008, \enquote{A ZMP Feedback
  Control for Biped Balance its Application to In-Place Lateral Stepping
  Motion.} J. Comput., \textbf{3}(8), pp. 23--31.

\bibitem{8461140}
Seyde, T., Shrivastava, A., Englsberger, J., Bertrand, S., Pratt, J., and
  Griffin, R.~J., 2018, \enquote{Inclusion of Angular Momentum During Planning
  for Capture Point Based Walking,} \textit{2018 IEEE International Conference
  on Robotics and Automation (ICRA)}, pp. 1791--1798.

\bibitem{7de9e7e7aa784c1e8f9fa707179496b7}
Morimoto, J., Endo, G., Nakanishi, J., and Cheng, G., 2008, \enquote{A
  biologically inspired biped locomotion strategy for humanoid robots:
  Modulation of sinusoidal patterns by a coupled oscillator model,} IEEE
  Transactions on Robotics, \textbf{24}(1), pp. 185--191.

\bibitem{ijmker2013energy}
Ijmker, T., Houdijk, H., Lamoth, C.~J., Beek, P.~J., and van~der Woude, L.~H.,
  2013, \enquote{Energy cost of balance control during walking decreases with
  external stabilizer stiffness independent of walking speed,} Journal of
  biomechanics, \textbf{46}(13), pp. 2109--2114.

\bibitem{donelan2004mechanical}
Donelan, J.~M., Shipman, D.~W., Kram, R., and Kuo, A.~D., 2004,
  \enquote{Mechanical and metabolic requirements for active lateral
  stabilization in human walking,} Journal of biomechanics, \textbf{37}(6), pp.
  827--835.

\bibitem{koopman2013lateral}
Koopman, B., Meuleman, J., van Asseldonk, E.~H., and van~der Kooij, H., 2013,
  \enquote{Lateral balance control for robotic gait training,} \textit{2013
  IEEE 13th International Conference on Rehabilitation Robotics (ICORR)}, IEEE,
  pp. 1--6.

\bibitem{bauby2000active}
Bauby, C.~E. and Kuo, A.~D., 2000, \enquote{Active control of lateral balance
  in human walking,} Journal of biomechanics, \textbf{33}(11), pp. 1433--1440.

\bibitem{perry2017walking}
Perry, J.~A. and Srinivasan, M., 2017, \enquote{Walking with wider steps
  changes foot placement control, increases kinematic variability and does not
  improve linear stability,} Royal Society open science, \textbf{4}(9), p.
  160627.

\bibitem{kuo1999stabilization}
Kuo, A.~D., 1999, \enquote{Stabilization of lateral motion in passive dynamic
  walking,} The International journal of robotics research, \textbf{18}(9), pp.
  917--930.

\bibitem{sullivan20143d}
Sullivan, T. and Seipel, J., 2014, \enquote{3D dynamics of bipedal running:
  Effects of step width on an amputee-inspired robot,} \textit{2014 IEEE/RSJ
  International Conference on Intelligent Robots and Systems}, IEEE, pp.
  939--944.

\bibitem{rezazadeh2020control}
Rezazadeh, S. and Hurst, J.~W., 2020, \enquote{Control of ATRIAS in three
  dimensions: Walking as a forced-oscillation problem,} The International
  Journal of Robotics Research, \textbf{39}(7), pp. 774--796.

\bibitem{rezazadeh2015spring}
Rezazadeh, S., Hubicki, C., Jones, M., Peekema, A., Van~Why, J., Abate, A., and
  Hurst, J., 2015, \enquote{Spring-mass walking with atrias in 3d: Robust gait
  control spanning zero to 4.3 kph on a heavily underactuated bipedal robot,}
  \textit{Dynamic Systems and Control Conference}, p. V001T04A003.

\bibitem{winter2009biomechanics}
Winter, D.~A., 2009, \textit{Biomechanics and motor control of human movement},
  John wiley \& sons.

\bibitem{semasinghe2025design}
Semasinghe, C., Taylor, D., and Rezazadeh, S., 2025, \enquote{The Design and
  Manufacturing of Mithra: A Humanoid Robot with Anthropomorphic Attributes and
  High-Performance Actuators,} Robotics, \textbf{14}(3), p.~28.

\bibitem{geyer2006compliant}
Geyer, H., Seyfarth, A., and Blickhan, R., 2006, \enquote{Compliant leg
  behaviour explains basic dynamics of walking and running,} Proceedings of the
  Royal Society B: Biological Sciences, \textbf{273}(1603), pp. 2861--2867.

\bibitem{shamaei2013estimationhip}
Shamaei, K., Sawicki, G.~S., and Dollar, A.~M., 2013, \enquote{Estimation of
  quasi-stiffness of the human hip in the stance phase of walking,} PloS one,
  \textbf{8}(12), p. e81841.

\bibitem{molitor2024lower}
Molitor, S.~L. and Neptune, R.~R., 2024, \enquote{Lower-limb joint
  quasi-stiffness in the frontal and sagittal planes during walking at
  different step widths,} Journal of Biomechanics, \textbf{162}, p. 111897.

\bibitem{shamaei2013estimation}
Shamaei, K., Sawicki, G.~S., and Dollar, A.~M., 2013, \enquote{Estimation of
  quasi-stiffness and propulsive work of the human ankle in the stance phase of
  walking,} PloS one, \textbf{8}(3), p. e59935.

\bibitem{arampatzis1999effect}
Arampatzis, A., Br{\"u}ggemann, G.-P., and Metzler, V., 1999, \enquote{The
  effect of speed on leg stiffness and joint kinetics in human running,}
  Journal of biomechanics, \textbf{32}(12), pp. 1349--1353.

\end{thebibliography}



\end{document}